\documentclass[lettersize,journal]{IEEEtran}
\usepackage{amsmath,amsfonts}
\usepackage{array}
\usepackage{textcomp}
\usepackage{stfloats}
\usepackage{url}
\usepackage{verbatim}
\usepackage{graphicx}
\def\BibTeX{{\rm B\kern-.05em{\sc i\kern-.025em b}\kern-.08em
    T\kern-.1667em\lower.7ex\hbox{E}\kern-.125emX}}
\usepackage{balance}
\usepackage{subcaption}

\usepackage{soul}
\usepackage{xcolor}
\sethlcolor{yellow}

\usepackage{graphics} 
\usepackage{epsfig} 
\usepackage{amsmath} 
\usepackage{amssymb}  
\usepackage[noend]{algpseudocode}
\usepackage{bm}
\usepackage{multirow}
\usepackage{booktabs}
\usepackage{float}
\usepackage{placeins}
\usepackage{subcaption}
\usepackage{url}

\usepackage{flushend}
\usepackage{hyperref}

\usepackage{xspace}
\makeatletter
\DeclareRobustCommand\onedot{\futurelet\@let@token\@onedot}
\def\@onedot{\ifx\@let@token.\else.\null\fi\xspace}

\def\eg{\emph{e.g}\onedot} 
\def\ie{\emph{i.e}\onedot}

\def\etal{\emph{et al}\onedot}
\makeatother

\DeclareMathOperator*{\argmin}{arg\,min}

\newcommand{\bzero}{\mathbf{0}}

\newcommand{\bA}{\mathbf{A}}

\newcommand{\cD}{\mathcal{D}}

\newcommand{\cL}{\mathcal{L}}

\newcommand{\cE}{\mathcal{E}}

\newcommand{\cS}{\mathcal{S}}
\newcommand{\bI}{\mathbf{I}}

\newcommand{\bbI}{\mathbb{I}}

\newcommand{\br}{\mathbf{r}}
\newcommand{\bP}{\mathbf{P}}
\newcommand{\bp}{\mathbf{p}}
\newcommand{\bQ}{\mathbf{Q}}

\newcommand{\bR}{\mathbf{R}}

\newcommand{\bbR}{\mathbb{R}}
\newcommand{\bbZ}{\mathbb{Z}}

\newcommand{\bt}{\mathbf{t}}

\newcommand{\bX}{\mathbf{X}}
\newcommand{\bY}{\mathbf{Y}}
\newcommand{\bx}{\mathbf{x}}

\newcommand{\be}{\mathbf{e}}
\newcommand{\bz}{\mathbf{z}}

\newcommand{\bK}{\mathbf{K}}

\newcommand{\btheta}{\boldsymbol\theta }

\newcommand{\bomega}{\bm{\omega}}
\newcommand{\bSigma}{\bm{\Sigma}}
\newcommand{\bPi}{\bm{\Pi}}
\newcommand{\bchi}{\bm{\chi}}
\newcommand{\bsigma}{\bm{\sigma}}
\newcommand{\btau}{\bm{\tau}}

\newcommand{\neuropnp}{NeuroPnP}

\newcommand{\snnlandmarkregressor}{SPose}

\begin{document}
\title{Robust PnP on a Neuromorphic Processor for Object Pose Estimation}
\author{Tam Ngoc-Bang Nguyen, Mohsi Jawaid, Tat-Jun Chin
\thanks{All authors are with the AI for Space Group, Adelaide University, Australia. Corresponding author: T.N-B.Nguyen. Email address: \href{mailto:tam.nb.nguyen@adelaide.edu.au}{tam.nb.nguyen@adelaide.edu.au}.}}

\markboth{Journal of \LaTeX\ Class Files,~Vol.~18, No.~9, September~2020}%
{How to Use the IEEEtran \LaTeX \ Templates}

\maketitle

\begin{abstract}
Neuromorphic computing is gaining attention in robotic perception due to its higher energy efficiency. While neural network-based methods can more readily exploit the distributed and parallelized structure of neuromorphic computers, crafting neuromorphic solutions for non-learning tasks is less straightforward. This hampers the usage of neuromorphic computing for perception pipelines that depend on both learning and non-learning components, such as object pose estimation (OPE) where state-of-the-art methods use a deep network to predict 2D landmarks and nonlinear optimization to solve perspective-n-point (PnP). In this paper, we propose a novel neuromorphic-deployable formulation for robust PnP, where given outlier-prone 2D-3D correspondences, the object pose with the largest number of inliers is determined. Underpinning our method is a distributed algorithm for robust least squares estimation of rigid body pose that can be executed on a neuromorphic processor. We also design a spiking neural network (SNN) to predict 2D landmarks from event data, where the main layers of the SNN were designed according to the principles of spiking neurons. Overall, our work enables neuromorphic treatment of the major stages of an OPE pipeline, from event sensing and learned landmark prediction, to geometric optimization for robust PnP\footnote{Our demo program will be released if accepted.}. Results on neuromophic hardware (Intel Loihi~2) indicate the higher energy efficiency our neuromorphic robust PnP, while achieving competitive accuracy.
\end{abstract}


\section{Introduction}

Object pose estimation (OPE) is a fundamental perception capability that enables embodied AI systems to interact with objects in their physical environment~\cite{collet2011moped, geiger2012we, chen2017multi, xu2018pointfusion, zimpfer2005autonomous, opromolla2017review}. Generally speaking, the goal of OPE is estimating the 6-DoF pose of a target object relative to the observer. In this work, we focus on model-based OPE for rigid (non-articulated) objects, where a known 3D model of the object is assumed available.

In the chosen setting, an effective approach involves predicting the 2D locations of pre-determined points on the 3D model using a learned 2D landmark regressor~\cite{rad2017bb8, gupta2019cullnet, song2020hybridpose, su2022zebrapose,ausserlechner2024zs6d, xu2024rnnpose}, which yields a 2D-3D correspondence set, then solving the perspective-n-point (PnP) problem via nonlinear optimization to recover the pose~\cite{lu2018review, pan2021survey}. Such two-stage OPE methods are often more accurate than end-to-end learning approaches that directly predict the pose~\cite{liu2025picopose, wang2025hccepose}.

The increasing emphasis on deploying embodied AI systems in the real world demands perception capabilities that can be executed on edge compute devices. However, such devices are usually subject to stringent size, weight and power (SWaP) constraints. In particular, since an untethered embodied AI platform must operate on battery power, an edge device that is energy inefficient will quickly drain the power and limit the uptime and responsiveness of the embodied AI.

Neuromorphic computing~\cite{schuman2022opportunities} has emerged as a promising approach to energy-efficient AI. Inspired by biological neural architectures, a neuromorphic computer comprises a distributed and parallel network of compute nodes interconnected via synapses whose weights determine the connection strengths. Each node evolves according to its internal temporal dynamics and communicates asynchronously with others through spike-based messages. Together, the network architecture, synaptic weights, and neuronal dynamics constitute a neuromorphic algorithm. When implemented on specialized neuromorphic processors~\cite{poon2011neuromorphic, merolla2014million, orchard2021efficient, davies2018loihi, davies2021advancing}, such algorithms typically draw less power or consume less energy than equivalent algorithms running on conventional machines.

Existing efforts in neuromorphic algorithm development can be broadly grouped into learning and non-learning approaches. Works in the former often invoke the spiking neural network (SNN) formalism~\cite{maass1997networks} and learn the synapse weights from training data while keeping each node as a relatively simple spiking neuron~\cite{eshraghian2023training}. Methods in the latter tend to directly design the neuromorphic algorithm, including the architecture and node dynamics~\cite{aimone2022review}, usually to solve non-cognitive tasks such as combinatorial optimization.

\begin{figure*}[t]
    \centering
    \includegraphics[width=\textwidth]{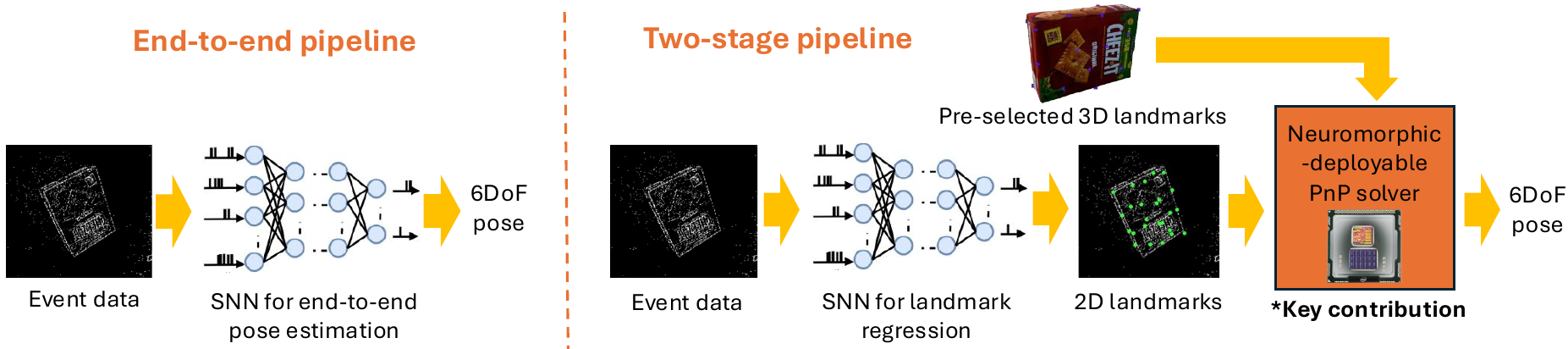}
    \caption{Contrasting end-to-end (left) and two-stage (right) neuromorphic OPE pipelines. An end-to-end method trains an SNN to directly predict the 6DoF pose of an object captured in the event data. A two-stage method trains an SNN to regress 2D landmarks on the object first, before solving PnP on the 2D-3D correspondences to estimate the object pose. Our key contribution is a distributed algorithm for robust PnP that can be executed on a neuromorphic processor. Combined with a novel SNN for landmark regression, our work enables two-stage OPE that is compatible with neuromorphic hardware.}
    \label{fig:wide_figure}
\end{figure*}

In embodied AI, neuromorphic computing has demonstrated significant gains in energy efficiency while achieving competitive performance across a range of applications, including perception, control, and navigation tasks~\cite{ziegler2025detection, mangalore2024neuromorphic, hines2025compact, renner2024visual, tang2019spiking, vitale2021event, shang2025fully, paredes2024fully, stroobants2025neuromorphic}. Despite this progress, neuromorphic OPE remains relatively underexplored, with only a few recent works addressing the problem~\cite{courtois2024spiking, rathinam2026efficient}. Courtois \etal~\cite{courtois2024spiking} developed a direct regression OPE pipeline, where a convolutional SNN was trained to regress 6-DoF poses on the SEENIC dataset~\cite{jawaid2023towards}. However, its pose error is approximately two times higher than that of conventional neural network baselines, and no neuromorphic hardware deployment or power measurements were reported. In contrast, Rathinam \etal~\cite{rathinam2026efficient} adopted a two-stage hybrid learning-classical pipeline, where a keypoint regressor SNN is deployed on the Akida neuromorphic processor. Nevertheless, the geometric PnP solver remains executed on a CPU. In short, a fully neuromorphic OPE pipeline is still missing. Fig.~\ref{fig:wide_figure} contrasts end-to-end and two-stage neuromorphic OPE.

To enable a comprehensive neuromorphic treatment of OPE, it is vital to be able to solve non-learning components such as PnP within the neuromorphic framework. This is non-trivial as geometric vision tasks typically demand constrained optimization of nonlinear models (\eg, 6DoF pose in PnP) and involve outlier-prone input data (\eg, incorrect 2D-3D matches). Unsurprisingly, neuromorphic solutions to geometric vision have received little attention thus far.

\subsection{Contributions}

Our key contribution is a distributed formulation for robust PnP that is \emph{neuromorphic deployable}, \ie, the algorithm can be structured as an SNN and defined using the instruction set of a neuromorphic processor (Intel Loihi 2). Although the proposed algorithm is not strictly neuromorphic (\eg, it does not rely on asynchronous spike processing), it can exploit the hardware to conduct robust PnP with high energy efficiency.

To complement the primary contribution above, we also developed an SNN for 2D landmark regression on event data. The main layers of the SNN were designed under the constraints of a real neuromorphic processor, while the overall SNN can be trained using the Spiking Jelly framework~\cite{fang2023spikingjelly}.

Combined, our contributions realize a neuromorphic OPE pipeline where the major components of sensing, landmark prediction and robust PnP are all amenable to neuromorphic hardware execution; see Fig.~\ref{fig:wide_figure}~(right). Results show that our pipeline is as accurate as equivalent classical pipelines.

\section{Related work}

\subsection{SNNs in computer vision}

Many works approach neuromorphic computing in computer vision via the SNN route, where SNNs are either trained directly using surrogate-gradient backpropagation or converted from pre-trained artificial neural networks (ANNs)~\cite{eshraghian2023training}. In recent years, SNNs have demonstrated significant gains in energy efficiency while achieving competitive performance across diverse applications, including image classification~\cite{fang2021deep, hu2024advancing}, segmentation~\cite{patel2021spiking, lei2025spike2former}, object detection~\cite{kim2020spiking, cordone2022object, su2023deep, luo2024integer, li2025brain, ahmed2025efficient}, and optical flow estimation~\cite{schnider2023neuromorphic}. Extensive benchmarking of SNNs implemented on numerous neuromorphic processors, such as IBM TrueNorth~\cite{merolla2014million}, SpiNNaker~\cite{gonzalez2024spinnaker2}, Intel Loihi 1 and 2~\cite{davies2018loihi, davies2021advancing, orchard2021efficient}, further highlights their potential to reduce the energy consumption compared to conventional CPU- or ANN-based approaches. However, less effort has been directed toward non-learning vision tasks, which limits the development of neuromorphic pipelines that combine both learning and non-learning components.

\subsection{Neuromorphic solutions for non-cognitive problems} 

This research leverages the event-driven property and massive parallelism of the neuromorphic computing to address non-cognitive computational tasks such as constraint satisfaction~\cite{fonseca2017using, binas2016spiking, jonke2016solving, davies2021advancing}, combinatorial optimization~\cite{pierro2024solving, fang2022solving, corder2018solving, nguyen2024slack}, and continuous optimization~\cite{mangalore2024neuromorphic, tang2017sparse}. In these works, problem variables and constraints are embedded within the network’s topology and stateful dynamics, enabling the system to naturally evolve toward low-energy or constraint-satisfying solutions. By directly exploiting neuromorphic dynamics and parallelism, these methods offer a highly parallel and energy-efficient alternative to traditional optimization. Our work shares this spirit by rethinking the PnP problem within a neuromorphic framework to enable a low-power and energy-efficient solution.

\subsection{Model-based OPE}

A class of methods for instance-level model-based 6DoF OPE directly regress the 6DoF pose of the object using deep networks~\cite{xiang2017posecnn, capellen2019convposecnn, kendall2015convolutional}. Such approaches, however, are generally less accurate than two-stage pipelines~\cite{tekin2018real, peng2019pvnet, chen2022occlusion, su2022zebrapose, wang2025hccepose, liu2025picopose}. Generally in two-stage pipelines, 2D–3D correspondences are first established, either through classical feature matching or via a landmark regressor network; and the final pose is subsequently estimated using a PnP solver~\cite{lepetit2009ep, lu2018review, pan2021survey}. Owing to their superior accuracy and robustness across diverse settings~\cite{wang2025hccepose}, two-stage pipelines remain a prevalent direction for instance-level model-based OPE. 

Recently, neuromorphic sensing for OPE has attracted interest due their potentially higher robustness under challenging scenarios such as fast motion or extreme lighting conditions~\cite{glover2024edopt, liu2024line,jawaid2023towards, jawaid2024test, malik2025evsat3d}. However, most methods accumulate event data into intensity frames and apply conventional (non-neuromorphic) algorithms. An early work neuromorphic processing for event-based OPE is~\cite{courtois2024spiking}, which developed an SNN to predict 6DoF pose from input events. As alluded to above, direct pose regression is generally less accurate than two-stage methods. Also, Rathinam \etal~\cite{rathinam2026efficient} employed a hybrid solution for their two-stage pipeline, where a keypoint regressor SNN is deployed on the Akida neuromorphic processor, while PnP is executed on a classical CPU. The literature is missing a neuromorphic PnP solver to complete the picture.

\subsection{OPE datasets}

While numerous RGB-based OPE benchmark datasets exist, namely LINEMOD~\cite{hinterstoisser2012model}, T-LESS~\cite{hodan2017t} and YCB-V~\cite{xiang2017posecnn}, relatively a smaller number of event-based counterparts are available: they include YCB-EV~\cite{rojtberg2024ycb}, RGB-D-E~\cite{dubeau2020rgb} E-POSE~\cite{hay2025pose} and FRESH~\cite{jawaid2025event}. To demonstrate our neuromorphic PnP solver, we will integrate it into a two-stage event-based OPE pipeline and evaluate it on the latter two datasets. Notably, E-POSE includes objects from YCB-EV, while FRESH provides higher-resolution event data of satellite-like objects captured under challenging lighting conditions.

\subsection{PnP solvers}

Extensive research has been devoted to improving PnP algorithms for computer vision~\cite{ding2023revisiting,lepetit2009ep,madsen2004methods, levenberg1944method}. Learning-based methods integrate differentiable PnP layers into end-to-end trainable pipelines~\cite{chen2022epro, chen2020end} to exploit rich training data to improve robustness in complex scenes. More recently, energy-efficient realizations of PnP have also been explored, including FPGA-based implementations that balance accuracy with low-power operation~\cite{sufi2024fpga, lv2024energy}. However, a neuromorphic treatment of PnP has received little attention.

\section{Preliminaries}

We first outline the basic ideas that our method is built upon. The presentation here is not meant to be comprehensive; please consult the references provided for details.

\subsection{PnP for OPE}\label{sec:pnpintro}

Let $\cD = \{ (\bx_i, \bX_i) \}^{N}_{i=1}$ be a set of 2D-3D correspondences, where $\{ \bX_i \} \subset \mathbb{R}^3$ are 3D points  in world frame that define the target object's shape, and $\{ \bx_i \} \subset \mathbb{R}^2$ are the their noisy 2D observations in an image $I$. PnP aims to recover the rigid transformation $(\bR,\bt) \in SE(3)$ that \emph{posed} $\{ \bX_i \}$ in the camera frame before being \emph{imaged} to become $\{ \bx_i \}$.

\paragraph{Nonlinear least squares (NLLS)} 

NLLS solves
\begin{equation}\label{eq:pnp_nnls}
    \min \limits_{(\bR,\bt) \in SE(3)} \sum^{N}_{i=1} \lVert \bx_i - f(\bX_i \mid \bP) \rVert_2^2,
\end{equation}
where $f$ is the posing and imaging process given by
\begin{align}
    f(\bX_i \mid \bP) = \frac{\bP^{(1:2)}\tilde{\bX}_i}{\bP^{(3)}\tilde{\bX}_i},
\end{align}
with $\widetilde{\bX}_i = [\bX_i^T \quad 1]^T$ being the homogeneous form of $\bX_i$,
\begin{align}
    \bP = \bK\left[ \begin{matrix}\bR & \bt \end{matrix} \right] \in \mathbb{R}^{3\times 4}
\end{align}
being the camera matrix, $\bK \in \mathbb{R}^{3\times 3}$ being the known camera intrinsics, and $\bP^{(1:2)}$ and  $\bP^{(3)}$ respectively being the first-two rows and third row of $\bP$. Effective NLLS algorithms~\cite{madsen2004methods, eade2013gauss} are available to solve~\eqref{eq:pnp_nnls}.

\paragraph{Direct linear transformation (DLT)} DLT solves
\begin{equation}\label{eq:LS_DLT}
    \bp^* = \argmin \limits_{\lVert \bp \rVert_2 = M} \lVert \bA \bp \rVert_2^2,
\end{equation}
where $\bp = \operatorname{vec}(\bP) \in \mathbb{R}^{12}$ is vectorisation of $\bP$, and $\bA \in \mathbb{R}^{2N \times 12}$ is obtained by vertically stacking the matrices
\begin{align}\label{eq:monomials}
    \bA_i = \begin{bmatrix}
        \bzero_{1 \times 4}^T & -\widetilde{\bX}_i & v_i \widetilde{\bX}_i^T \\
        \widetilde{\bX}_i^T & \bzero_{1 \times 4}^T & -u_i \widetilde{\bX}_i
    \end{bmatrix} \in \bbR^{2 \times 12} \quad \forall i,
\end{align}
where $(u_i,v_i)$ are the values of $\bx_i$. The norm constraint avoids the trivial solution $\bp = \bzero$. While $M=1$ is typically used, any $M \in \mathbb{R}_+$ is viable since $\bp$ is scale-invariant.

Since $\bp^\ast$ is the least significant right singular vector of $\bA$ (rescaled post hoc by $M$), the singular value decomposition (SVD) is usually employed for DLT~\cite[Alg.~4.1]{hartley2003multiple}. Given $\bp^\ast$, the pose can be recovered through solving the Procrustes problem~\cite{Schönemann_1966} as postprocessing. For the full derivation of DLT for PnP, refer to~\cite{henry2025optimaldltbasedsolutionsperspectivenpoint}.

DLT can be unstable when the magnitudes of the values in $\bA$ differ greatly due to the multiplications in~\eqref{eq:monomials}. This concern is conclusively addressed by appropriate preconditioning of $\cD$ and/or $\bA$, which can be conducted prior to invoking the solver; see~\cite[Sec.~4.4]{hartley2003multiple} and \cite[Chap.~2]{Kanatani2016}.

\paragraph{Robust estimation}
If $\cD$ contains outliers, conducting NLLS or DLT directly will lead to biased results. To estimate the pose robustly, the solver is embedded in a hypothesize-and-verify loop~\cite{fischler1981random}, where in each iteration a minimal subset of correspondences is randomly sampled to generate a pose hypothesis. The hypothesis with the largest number of inliers is returned as the result.

\subsection{Intel Loihi 2 neuromorphic processor}\label{sec:loihi2intro}

\begin{figure}[ht]\centering
\includegraphics[width=0.99\columnwidth]{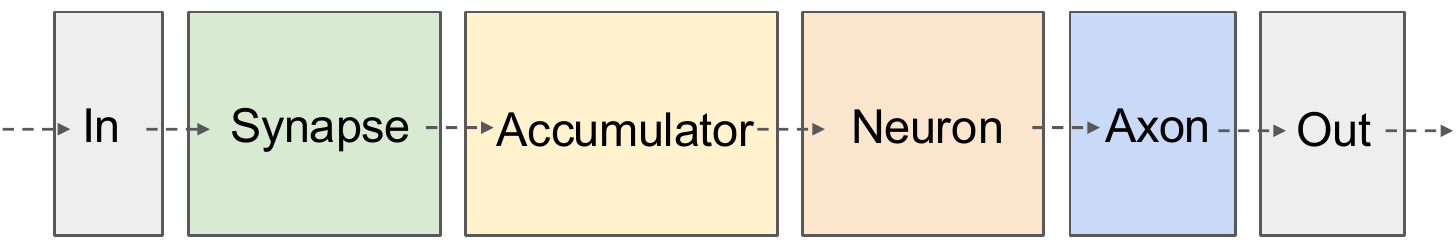}
\caption{Basic schematic of a neuromorphic core in Intel Loihi 2. The Neuron and Synapse are programmable/configurable.}
\label{fig:loihi2}
\end{figure}

Basic information on Intel Loihi~2~\cite{davies2018loihi, orchard2021efficient} is provided, since the proposed PnP algorithm is targeted for this particular neuromorphic processor. The device is designed for event-driven and massively parallel computation. The chip integrates 128 neuromorphic cores, which are able to support 8192 logical Neurons, each programmable with 8-bit synaptic connections; Fig.~\ref{fig:loihi2} illustrates. Incoming messages from other neurons are processed by a \emph{Synapse} block, which performs matrix–vector multiplications or convolutions, and the resulting activations are accumulated and sent to the \emph{Neuron} block, where neuron programs of up to 24-bit internal states are executed. The interconnections and neuronal dynamics on Loihi 2 are user-definable through assembly-level programs. Note that despite the neural network-heavy terminology, a neuromorphic processor can also solve non-learning tasks~\cite{aimone2022review}.

\section{Neuromorphic-deployable robust PnP}

Here, we describe the proposed robust PnP algorithm that can be deployed on a neuromorphic processor. The overarching idea is constructing distributed versions of nonlinear geometric optimization and robust estimation, whose structure and update steps can then be respectively defined as an SNN and neuronal dynamics. Strictly speaking, while this does yield a neuromorphic PnP algorithm (\eg, our approach does not conduct spike-based asynchronous message passing), our algorithm can be deployed on real neuromorphic hardware and benefit from its energy efficiency, as we will show in Sec.~\ref{sec:results}.

\subsection{Neuromorphic-deployable PnP optimization}\label{sec:formulation}

The $SE(3)$ constraints in NLLS and the SVD of $\bA$ for DLT are unwieldy for a neuromorphic treatment. To surmount this obstacle, we adopted the inverse iteration method~\cite{ipsen1997computing} to solve DLT. In each iteration, the linear system
\begin{equation}
    \label{eq:inverse_iteration}
    (\bQ - \mu \bI) \bp_{k+1} = \bp_k
\end{equation}
is solved, where $\bp_k$ and $\bp_{k+1}$ are resp.~the current and next iterates, $\bQ = \bA^T \bA$, and $\mu \in \bbR$ is the shift parameter. Before proceeding to the next iteration, $\bp_{k+1}$ is renormalised
\begin{align}
    \bp_{k+1} = M\frac{\bp_{k+1}}{\| \bp_{k+1} \|_2}.
\end{align}
If $\mu$ is close to $0$ (we used $\mu = 0$ in our work), then $\bp_k$ converges to the DLT solution after finite ($K$) iterations.

Our primary insight is that each update step of inverse iteration is the constrained linear least squares problem
\begin{align}
    \label{eq:QP_inverse_iteration}
    \bp_{k+1} &= \argmin \limits_{\lVert \btheta \rVert_2 = M } \| (\bQ - \mu \bI) \btheta - \bp_k \|_2^2\\
    &= \argmin \limits_{\lVert \btheta \rVert_2 = M } \frac{1}{2} \btheta^T (\bQ - \mu \bI) \btheta - \bp_k^T \btheta.
\end{align}
More importantly, the minimisation of the quadratic cost can be viewed as an \emph{energy minimisation process} that aligns with the native ability of neuromorphic computers~\cite{mangalore2024neuromorphic, pierro2024solving, nguyen2024slack, theilman2025solving}. Specifically, we define the Lagrangian
\begin{equation}
    \label{eq:QP_inverse_iteration_lagrange}
    \cL(\btheta, \lambda) = \frac{1}{2} \btheta^T (\bQ - \mu \bI) \btheta - \bp_k^T \btheta + \lambda (\lVert \btheta \rVert^2_2 - M^2)
\end{equation}
that encodes the energy landscape, where the third term enforces the norm constraint via a Lagrange multiplier $\lambda$. Minimising $\cL$ yields the constrained solution $\btheta^*$, which corresponds to the equilibrium of the process.

To reach this equilibrium, we construct iterative update rules for $\btheta$ and $\lambda$ based on projected gradient descent, \ie,
\begin{align}\label{eq:updatesteps}
\begin{aligned}
    \btheta^{(t+1)} &= \btheta^{(t)} - \alpha \left( \left( \bQ - \mu \bI \right) \btheta^{(t)} - \bp_k + 2 \lambda^{(t)} \btheta^{(t)} \right),  \\ 
    \lambda^{(t+1)} &= \lambda^{(t)} + \beta \left( \lVert \btheta^{(t)} \rVert^2_2 - M^2 \right),
\end{aligned}    
\end{align}
where $\alpha$ is the step size and $\beta$ controls the dual update rate. Intuitively, updating $\lambda$ ensures that an appropriate penalty is imposed on violations to the norm constraint.

The update steps~\eqref{eq:updatesteps} are amenable to the neuromorphic fabric (Fig.~\ref{fig:loihi2}). We institute a neuron $\theta_d$, $d = 1,\dots,12$, for each element of $\btheta$ and a neuron for $\lambda$. Collecting $\bomega = \left[\btheta^T,\lambda \right]^T$, we can define the neuronal dynamics as
\begin{align}
    \theta_d &\leftarrow \theta_d - \underbrace{\bomega^T \bSigma_d \bomega - \bsigma_d \bomega}_{synapse} + \underbrace{\alpha p_{k,d}}_{bias},\label{eq:dynamics_theta}\\
    \lambda &\leftarrow \lambda + \underbrace{\bomega^T \bPi \bomega}_{synapse} - \underbrace{\beta M^2}_{bias},\label{eq:dynamics_lambda}
\end{align}
where $p_{k,d}$ is the $d$-th element of $\bp_k$, 
\begin{align}
    \bSigma_d &= \be_d \be_{13}^T + \be_{13} \be_{d}^T,
\end{align}
with $\be_d$ being a vector with 1 at the $d$-th element and 0 elsewhere, $\bsigma_d$ is the $d$-th row of
\begin{align}
&\bsigma = [\alpha \bQ - \alpha \mu \bI \,|\ \bzero ] \in \bbR^{12 \times 13},\\
\text{and} \;\;\;\;\;\; &\bPi = \begin{bmatrix}
    \bI_{12 \times 12} & \bzero_{12 \times 1} \\
    \bzero_{1 \times 12} & 0
\end{bmatrix} \in \bbR^{13 \times 13}
\end{align}
The matrix-vector multiplications required can be performed by the synapse block, \ie, in hardware.

Conceptually, the notation in~\eqref{eq:dynamics_theta} and~\eqref{eq:dynamics_lambda}, which avoids the usage of time steps, alludes to the fact that the neurons evolve independently and time-asynchronously to reach the equilibrium in the overall energy~\eqref{eq:QP_inverse_iteration_lagrange}. Once equilibrium is attained, the state of $\btheta$ installed as $\bp_{k+1}$. The energy minimisation process is conducted $K$ times to solve DLT.

\subsection{Neuromorphic-deployable robust estimation}\label{sec:robust}

\paragraph{Hypothesis sampling}

To enable minimal subset sampling~\cite{fischler1981random}, we introduce $N$ binary variables $\bz \in \{0,1\}^N$, where $z_i = 1$ indicates the selection of $(\bx_i,\bX_i)$. Note that for DLT~\eqref{eq:LS_DLT}, at least $6$ correspondences ($\equiv 12$ rows of $\bA$) are needed to instantiate $\bp$, thus each $z_i$ should turn on with probability $6/N$. While this does not guarantee selecting exactly $6$ correspondences, it is not debilitating since
\begin{itemize}
\item If fewer than $6$ were selected, the gradient becomes noisy and $\btheta$ merely suffers from a temporary divergence.
\item If more than $6$ were selected, the chance of obtaining all-inlier minimal subsets is lowered.
\end{itemize}
Both issues above can be alleviated by allowing more iterations $T$ in the hypothesise-and-verify loop.

However, using $\bz$ to select on-the-fly data that participates in the update is challenged by the fact that $\bA$ is already hardcoded in the synapse at runtime; see $\bsigma$ in~\eqref{eq:dynamics_theta}. To circumvent this issue, we introduce gating variables
\begin{equation}\label{eq:intermediate_layer}
    \btheta' = \bz \otimes \btheta = \left[ \theta'_{i,d} \right] \in \bbR^{12N},
\end{equation}
where $\otimes$ is the Kronecker product. Each $\theta'_{i,d} = z_i \theta_d$ for $i \in \{1,\dots,N \}$ and $d \in \{1,\dots,12\}$, with the dynamic
\begin{align}
    \theta'_{i,d} \leftarrow z_i \theta_d.
\end{align}
Intuitively, the gating variables ``gate'' the contribution of each correspondence based on the subset selection, and allows to modify the update rule for $\btheta$ as
\begin{equation}
    \btheta^{(t+1)} = \btheta^{(t)} - \alpha \left( \left( \bQ' - \mu \bI \right) \btheta'^{(t)} - \bp_k + 2 \lambda^{(t)} \btheta^{(t)} \right)
\end{equation}
where $\bQ' \in \mathbb{R}^{12 \times 12N}$ is the horizontal concatenation of
\begin{align}\label{eq:Q_prime}
    \bQ'_i = \left[\begin{smallmatrix} 
        \bY_i + v_i^2 \bY_i & 
        -u_i v_i \bY_i & 
        -u_i \bY_i \\
        -u_i v_i \bY_i &
        (i + u_i^2) \bY_i &
        -v_i \bY_i \\
        -u_i \bY_i &
        -v_i \bY_i &
        (v_i^2 +u_i^2) \bY_i  \\
        \end{smallmatrix}\right] \in \bbR^{12 \times 12}
\end{align}
for $i = 1,\dots,N$, and $\bY_i = \tilde{\bX}_i \tilde{\bX}_i^T$. Collecting $\btau = \left[ \btheta^T, \btheta^{\prime T}, \lambda \right]^T$, the dynamics~\eqref{eq:dynamics_theta} can be rewritten as
\begin{align}
    \theta_d &\leftarrow \theta_d - \underbrace{\btau^T \bSigma^\prime_d \btau - \bsigma^\prime_d \btau}_{synapse} + \underbrace{\alpha p_{k,d}}_{bias},\label{eq:dynamics_theta_exp}
\end{align}
where $\bsigma^\prime_d$ is the $d$-th row of
\begin{align}
\bsigma^\prime = [\alpha \bQ' - \alpha \mu \bI \,|\ \bzero_{12 \times 12N+ 1} ] \in \bbR^{12 \times (13+12N)}
\end{align}
and $\bSigma^\prime_d = \be_d \be_{13+12N}^T + \be_{13+12N}\be_d^T$. Note that the coefficients in the synapses remain frozen at runtime.

\paragraph{Hypothesis verification}
Given a pose hypothesis $\bp$, the remaining task is evaluating its consistency with the input correspondences. This requires computing the algebraic residuals and aggregating them per correspondence, \ie,
\begin{align}
    \br^\prime = \lVert \bA \bp \rVert_2^2, \;\;\;\; r_i = r^\prime_{2(i-1)+1}  + r^\prime_{2(i-1)+2}, \forall i, 
\end{align}
then thresholding the per-correspondence residuals and counting the number of inliers of $\bp$, \ie,
\begin{align}
    \Psi = \sum_{i=1}^N \bbI(r_i \leq \epsilon^2_{\text{in}}).
\end{align}
Since the associated dynamical equations are trivial, we do not list them here. Refer to the supp.~material for details.

\subsection{Overall algorithm: NeuroPnP}\label{sec:snn}

\begin{figure}[ht]\centering
\includegraphics[width=0.99\columnwidth]{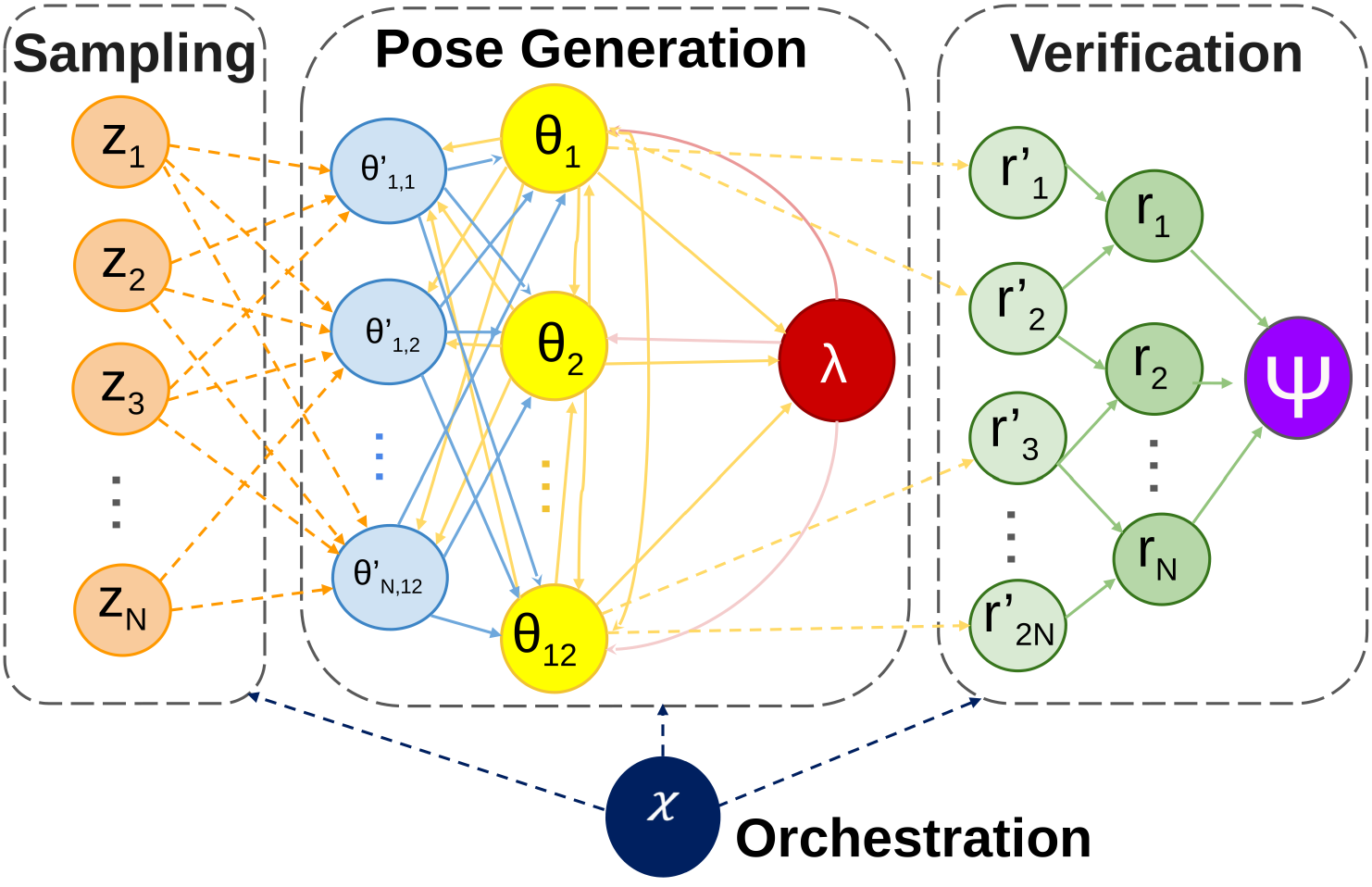}  
    \caption{Architecture of the proposed NeuroPnP algorithm. Solid and dotted lines resp.~indicate intra- and inter-module connections.}
    \label{fig:pnpsnn}
\end{figure}

Fig.~\ref{fig:pnpsnn} illustrates the proposed PnP algorithm, named \emph{NeuroPnP}, depicted as an SNN. The network solves~\eqref{eq:LS_DLT} robustly by repeatedly drawing a random subset of correspondences, generating pose hypotheses, and scoring them based on the number of inliers (Sec.~\ref{sec:robust}). Reflecting these main steps are the three main layers:
\begin{itemize}
    \item The \textit{Sampling} layer generates random subsets of 2D–3D correspondences for subsequent layers.
    \item Each selected subset is passed to the \textit{Pose Generation} layer, which estimates a pose hypothesis by solving a constrained energy minimisation problem.
    \item The resulting pose hypothesis is evaluated by the \textit{Verification} layer, which computes the number of inlier correspondences consistent with the estimated pose.
\end{itemize}
For pseudocode of the processing conducted in each neuron type, refer to the supplementary material.

When a neuromorphic program is active, the neurons operate independently. However, since NeuroPnP steps through three main distinct functions, some orchestration is required. To this end, we introduce a neuron $\bchi$ that is connected to all neurons to sequentially invoke/suppress the different layers. When a layer is suppressed, the neurons therein are inhibited from updating. To retrieve results from NeuroPnP, the states of the neurons are polled periodically.

\paragraph{Resource complexity}

Since neuromorphic cores integrate both computation and memory, the computational complexity of a neuromorphic algorithm can be quantified in terms of the number of neurons required~\cite{mangalore2024neuromorphic}. For NeuroPnP, the number of neurons required is 
\begin{align}\label{eq:scaling}
\begin{aligned}
    &\underbrace{N}_{Samp.} + \underbrace{12N + 12 + 1}_{Pose~Generation} + \underbrace{2N + N + 1}_{Verification} + \underbrace{1}_{Orches.}\\
    &= 16N + 15,
\end{aligned}
\end{align}
which is linear in the number $N$ of correspondences.

\subsection{Hyperparameter settings}\label{sec:hyperparam}

NeuroPnP's main hyperparameters and their setting are:
\begin{itemize}
    \item $\alpha$: step size for updating $\btheta = 0.005$
    \item $\beta$: step size for updating $\lambda = 0.1$
    \item $M$: norm $= 1$ for Lava simulation (64 bit CPU) and $10$ for Loihi 2 implementation
    \item $K$: number of inverse iterations $=50$ for Lava simulation (64 bit CPU) and $10$ for Loihi 2 implementation
    \item $G$: number of update steps for $\theta,\lambda = 1000$ for Lava simulation (64 bit CPU) and $50$ for Loihi 2 implementation
    \item $L$: number of hypothesise-and-verify loops $= 100$
    \item $\alpha\_exp$: scaling exponent of $\alpha = 10$, see Sec.~\ref{sec:hardware_details}
    \item $\beta\_exp$: scaling exponent of $\beta = 8$, see Sec.~\ref{sec:hardware_details}
\end{itemize}
The values were fixed throughout our experiments (Sec.~\ref{sec:results}). The inlier threshold $\epsilon_{\text{in}}$ was tuned for each input, but this parameter was shared with all competitor methods.

\subsection{Hardware implementation details}\label{sec:hardware_details}

As we will report in Sec.~\ref{sec:results}, the correctness of NeuroPnP has been verified via simulation using the Lava framework\footnote{\url{https://github.com/lava-nc/lava}}

We have also verified NeuroPnP on Loihi 2. Here, we provide an overview of the strategies employed to circumvent the limitations of the current hardware, including lower numerical precision (\eg, 8 bits in Synapse and 24 bits in Neuron), integer-only support, and a reduced instruction set without division operation~\cite{pierro2024solving, mangalore2024neuromorphic}. First, we adopted the fixed-point representation~\cite{mangalore2024neuromorphic} to approximate small floating-point numbers. For example, to represent $\alpha \in \bbR$, we convert it into a fixed-point counterpart
\begin{equation}
    \bar{\alpha} = \operatorname{ceil}(\alpha * 2^{\alpha\_exp}),
\end{equation}
where $\bar{\alpha}, \alpha\_exp \in \bbZ$ respectively denote the stored integer and its tuned scaling exponent. The division operation was emulated using right-shift bitwise approximations.

The strategies do not completely resolve the hardware constraints and noticeable approximation errors are inevitable. However, this reflects the limitation of the current hardware and not gaps in our algorithm.

\section{SNN for landmark regression}\label{sec:spose}

To illustrate the applicability of NeuroPnP in a two-stage neuromorphic OPE pipeline (Fig.~\ref{fig:wide_figure}), we also developed an SNN for 2D landmark regression from event data. We emphasize that the SNN presented here mainly plays a supporting role to our core contribution of NeuroPnP, hence no claims are made regarding its superiority over alternatives.

\subsection{Event data and representation}\label{sec:eventrepresentation}

\begin{figure}[t]\centering
\includegraphics[width=0.99\columnwidth]{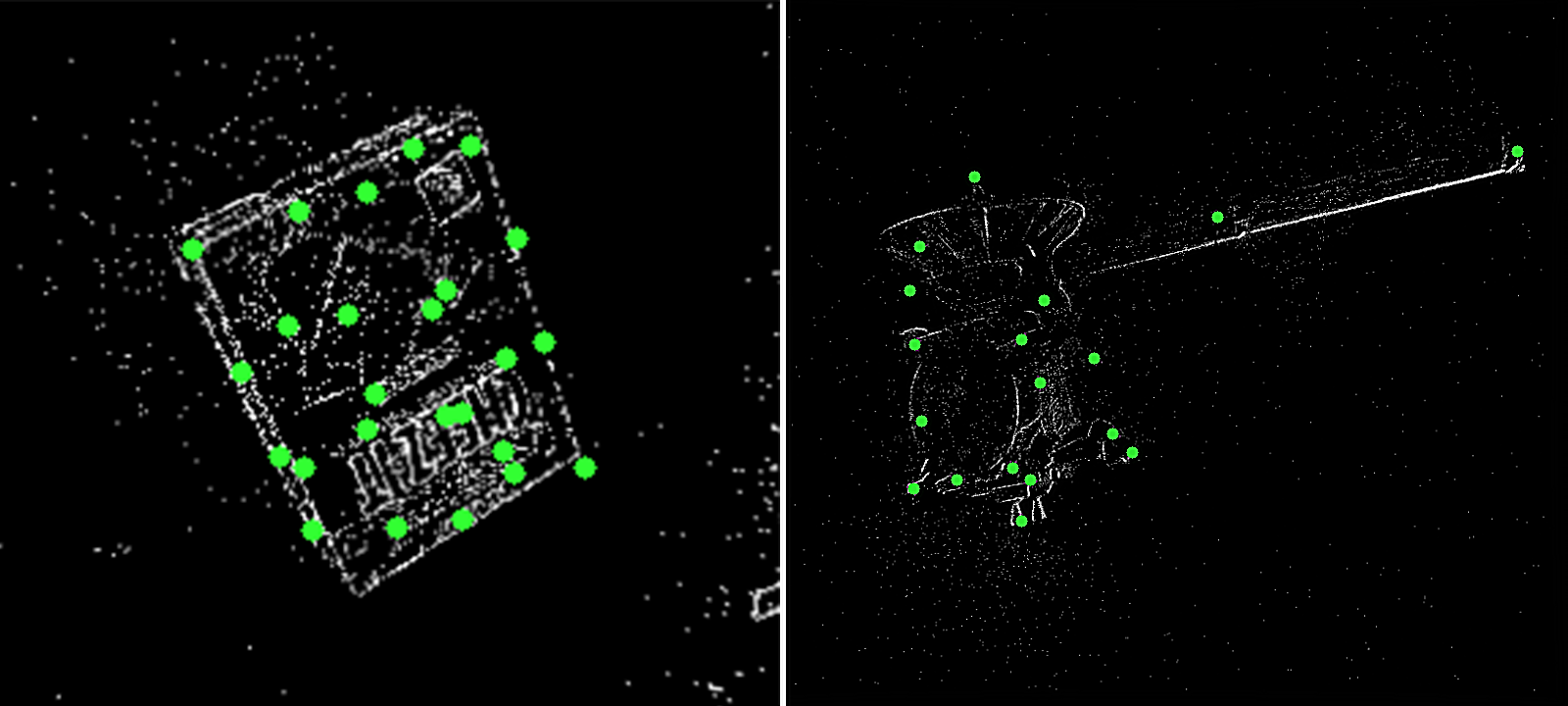}  
\caption{Example event frames with ground-truth landmarks from E-POSE~\cite{hay2025pose} (left) and FRESH~\cite{jawaid2025event} (right)}
\label{fig:EPOSE_FRESH}
\end{figure}

We assume that an event camera has recorded an event sequence $\cE$ containing the target object in the FOV of the camera. To derive an intermediate representation of $\cE$ for SNN inference, we partition $\cE$ into chunks of $\gamma$ ms and accumulate each chunk into an event frame. If the total duration of $\cE$ is $\delta$ ms, this creates a stack of $\delta/\gamma$ event frames. For consistency across the different datasets used (Sec.~\ref{sec:datasets}), we take $\delta = 50$ ms and $\gamma = 10$ ms; see Fig.~\ref{fig:EPOSE_FRESH} for  examples.

\subsection{SPose}\label{sec:landmark_regression}

Let $\cS = \{ e_j \}^{J}_{j=1}$ be a stack of sequential event frames, where each $e_j \in \mathbb{R}^{H \times W}$. For simplicity, we assume that a single instance of the target object is in $\cS$ with minor background clutter, and that the object pose is largely static within $\cS$. Our goal is predicting the 2D positions $\{ \bx_i \}^{N}_{i=1}$ of a set of known 3D landmarks $\{ \bX_i \}^{N}_{i=1}$ from $\cS$.

Each training event stack is paired with a set of ground truth 2D landmarks $\{\bx^*_i\}_{i=1}^N$, which we use to supervise an SNN-based landmark regressor. During inference, the trained SNN predicts 2D landmarks for unseen event stacks.

\begin{figure}[t]\centering
\includegraphics[width=\columnwidth]{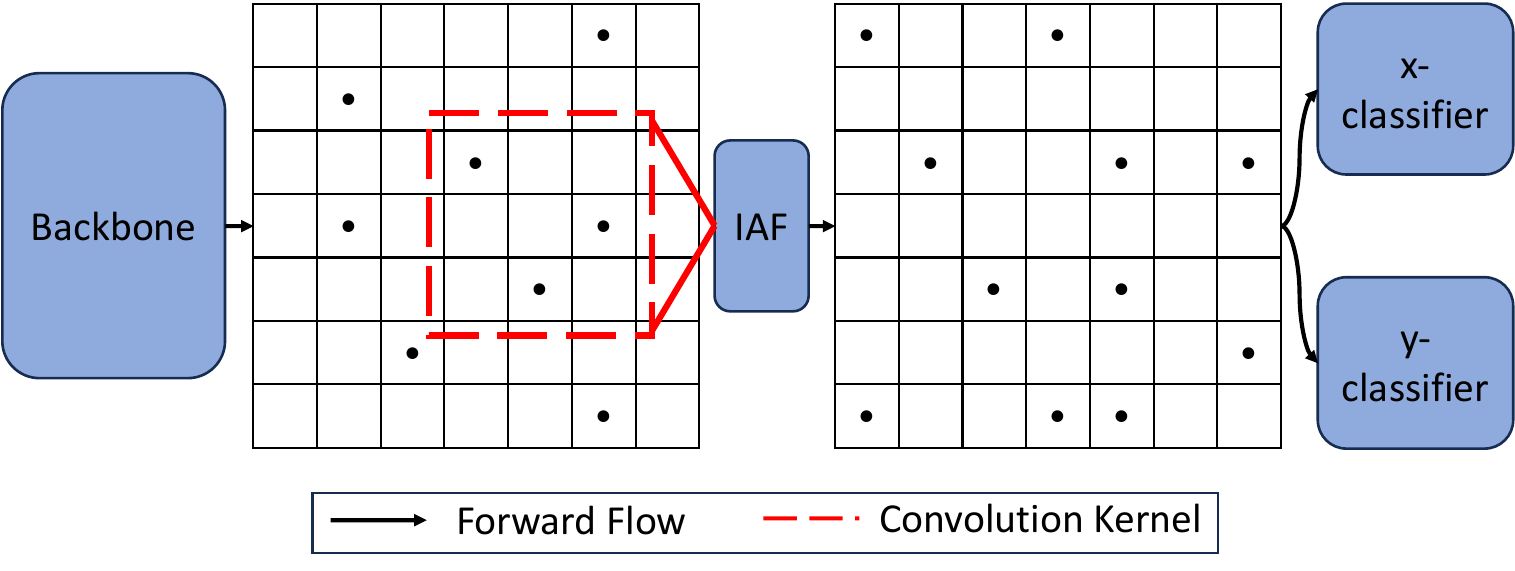}  
\caption{SPose architecture. Each grid represents a binary feature map, where dots indicate active cells.}
\label{fig:spose}
\end{figure}


The proposed SNN, called \emph{SPose}, is depicted in Fig.~\ref{fig:spose}. SPose is a spiking variant of SimCC~\cite{li2022simcc}, which originally comprises a ResNet-50 backbone, a convolutional layer, and two heads for x- and y-coordinate classification. To convert SimCC, the backbone is replaced with SEW-ResNet-50~\cite{fang2021deep}, and all ReLU activations are replaced with Integrate-and-Fire (IAF) neurons. An IAF layer is additionally inserted after the final convolutional layer to ensure the network is fully spiking prior to the classification heads. All convolutional layers are trained with 8-bit quantization-aware training to account for the lower precision of the current hardware.

Following SimCC~\cite{li2022simcc}, we represented each coordinate as a one-hot classification target over a discretized bin space, and directly trained the SNN using surrogate gradient methods~\cite{fang2023spikingjelly} with a Sigmoid surrogate function to minimize the cross-entropy between the ground-truth and predicted x- and y- coordinate bins for each landmark $\bx_i$ across each event stack $\cS$. The network outputs x- and y- coordinate bin distributions for each of the $N$ landmarks across $J$ event frames; we take the predictions from the last event frame as the final landmark estimates. Intuitively, we assume that the SNN integrates information over the entire event window, making the final frame the most informed estimate. These predictions yield the correspondence set $\cD = \{(\bx_i, \bX_i) \}^{N}_{i=1}$ for PnP.

\subsection{Hyperparameter settings}\label{sec:hyperparam_spose}

The hyperparameters for SPose were tuned for the datasets employed in our experiments (Sec.~\ref{sec:datasets}). Their values are
\begin{itemize}
\item J: stack size of sequential event frames $= 5$.
\item N: no.~of 3D landmarks = up to $25$ depending on object.
\item No.~of layers in SEW-ResNet-50 backbone: 56 (1 stem, 48 convolutions across 16 bottleneck blocks, 4 downsampling convolutions, 1 final convolution, 2 heads).
\end{itemize}
The settings above result in SNNs with up to $\approx 12.6$ Million spiking neurons, which unfortunately exceed the capacity of Intel Loihi 2. We therefore rely on CPU-based simulation of SPose in our experiments. Note that this reflects the current status of neuromorphic technology, rather than the shortcomings of our SNN architecture.

\section{Experiments}\label{sec:results}

Experiments were designed to evaluate the correctness of NeuroPnP, its accuracy relative to established PnP solutions, its performance on a real neuromorphic chip, and its applicability in a two-stage neuromorphic OPE pipeline (Fig.~\ref{fig:wide_figure}) with SPose as the SNN-based landmark regressor.

\subsection{Datasets and preprocessing}\label{sec:datasets}

We converted event sequences for the 13 objects from E-POSE~\cite{hay2025pose} and 3 objects from FRESH~\cite{jawaid2025event} into $5$-frame event stacks, following Sec.~\ref{sec:spose}. Ground-truth object poses at high frequency were available from the datasets, which we also used to generate ground-truth 2D landmarks based on up to 25 preselected 3D points on each CAD model.

We discarded event stacks where the object was not visible (\eg, due to occlusions or slow motion). This was achieved by setting an event count threshold and ignoring the chunks/stacks with a number of events less than the threshold. Overall, our preprocessing created 8295 event-based OPE/PnP input instances; see Fig.~\ref{fig:EPOSE_FRESH} for examples.

\subsection{Correctness of ~\neuropnp}\label{sec:correctness}

\paragraph{Methods}

To verify the correctness of NeuroPnP, we simulated it in the Intel Lava framework with 64-bit floating point precision and an x86-64 instruction set. The simulation was conducted on a standard machine with an Intel Core i7-11700K. Simulating NeuroPnP on a standard computer allows us to factor out the inadequacies of current neuromorphic hardware from the algorithm evaluation.

We compared NeuroPnP against
\begin{itemize}
    \item \texttt{RobustPnP-LM}: PnP with RANSAC as implemented on OpenCV\footnotemark{} with DLT and LM as the minimal solver.
    \item \texttt{RobustPnP-EPnP}: PnP with RANSAC as available on OpenCV\footnotemark[\value{footnote}] with EPnP~\cite{lepetit2009ep} as the minimal solver.
    \footnotetext{\url{https://docs.opencv.org/4.x/}}

\end{itemize}
For all the methods compared, $L =100$ hypothesise-and-verify iterations were given, and the pose returned by the robust estimator was refined by EPnP.

\begin{figure*}[t]\centering
\begin{subfigure}[b]{0.32\textwidth}
    \includegraphics[width=0.99\columnwidth]{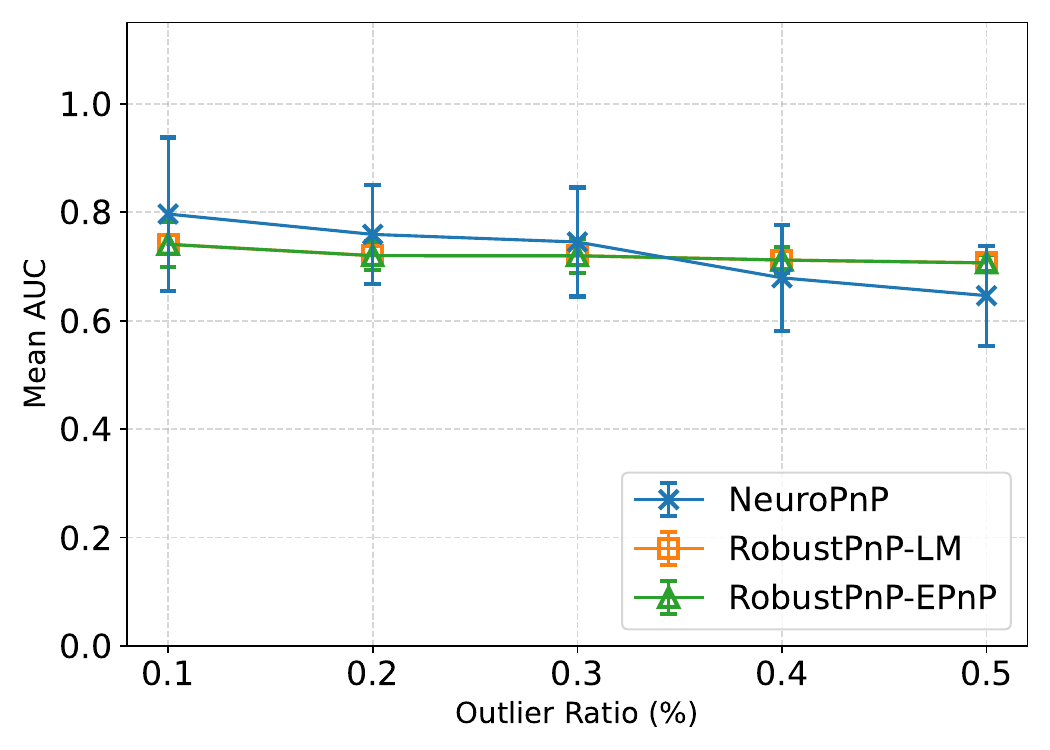}
    \caption{}
    \label{fig:epose_mean_AUC}
\end{subfigure}
\begin{subfigure}[b]{0.32\textwidth}
    \includegraphics[width=0.99\columnwidth]{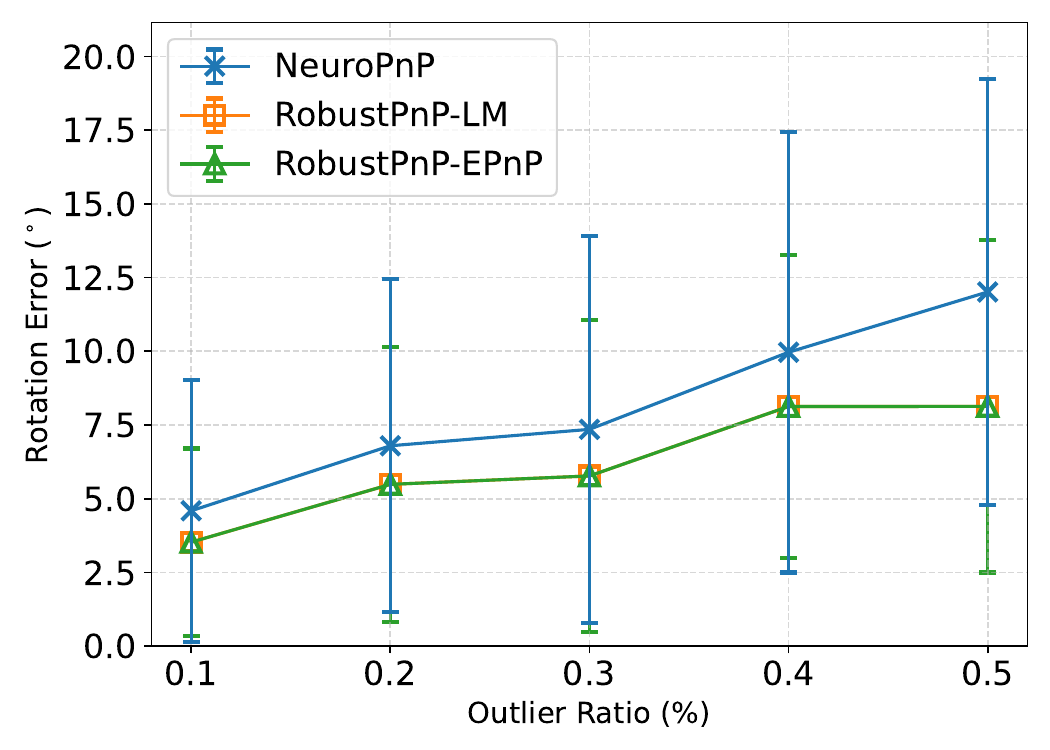}
    \caption{}
    \label{fig:epose_mean_Rot_err}
\end{subfigure}
\begin{subfigure}[b]{0.32\textwidth}
    \includegraphics[width=0.99\columnwidth]{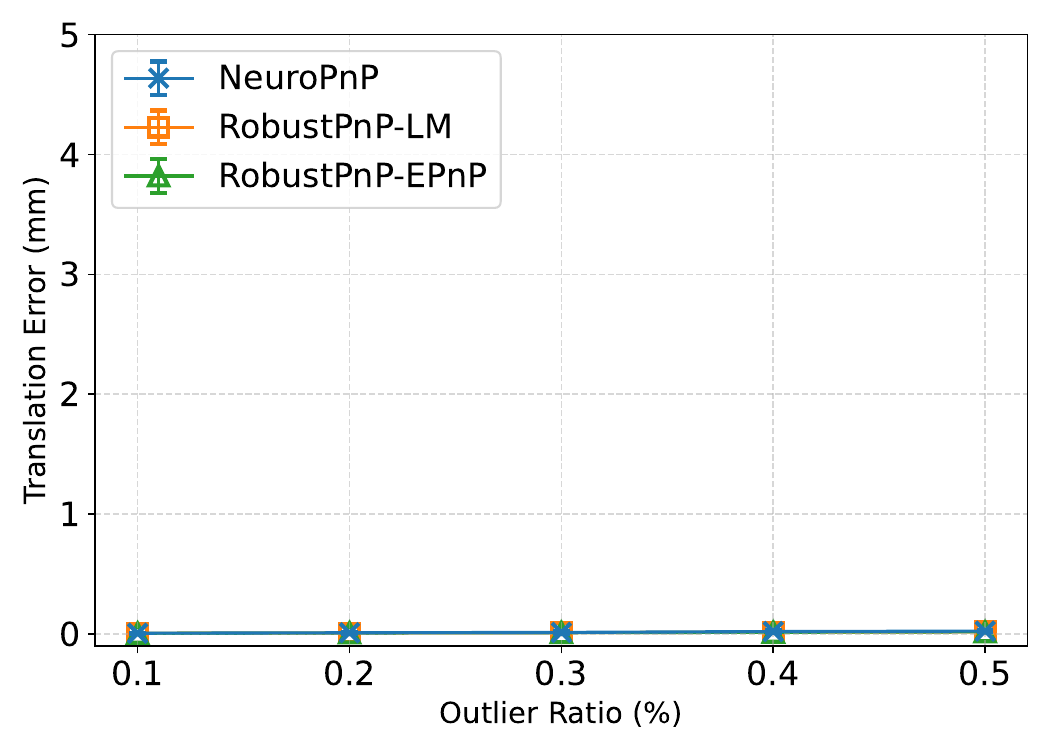}
    \caption{}
    \label{fig:epose_mean_Trans_err}
\end{subfigure}
\caption{Performance comparison between \neuropnp~and two robust PnP baselines across different outlier ratios on E-POSE. AUC values closer to 1 indicate superior inlier detection; lower rotation and translation errors indicate more accurate pose estimation.}
\label{fig:epose_verify_neuropnp}
\end{figure*}

\begin{figure*}[t]\centering
\begin{subfigure}[b]{0.32\textwidth}
    \includegraphics[width=0.99\columnwidth]{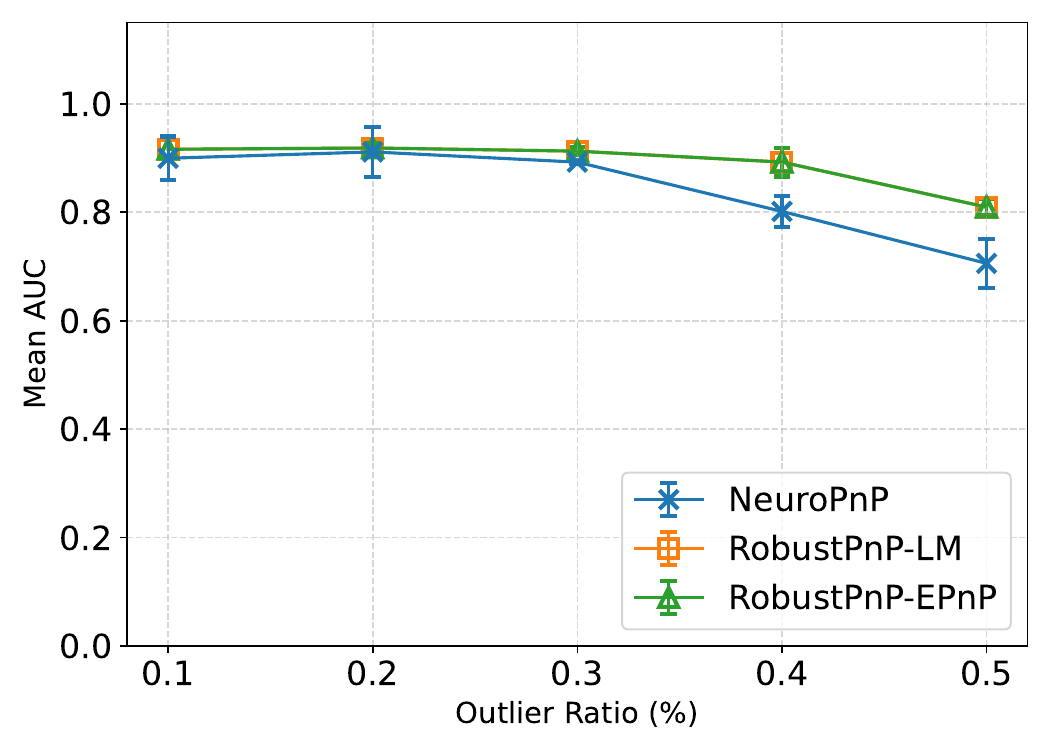}
    \caption{}
    \label{fig:fresh_mean_AUC}
\end{subfigure}
\begin{subfigure}[b]{0.32\textwidth}
    \includegraphics[width=0.99\columnwidth]{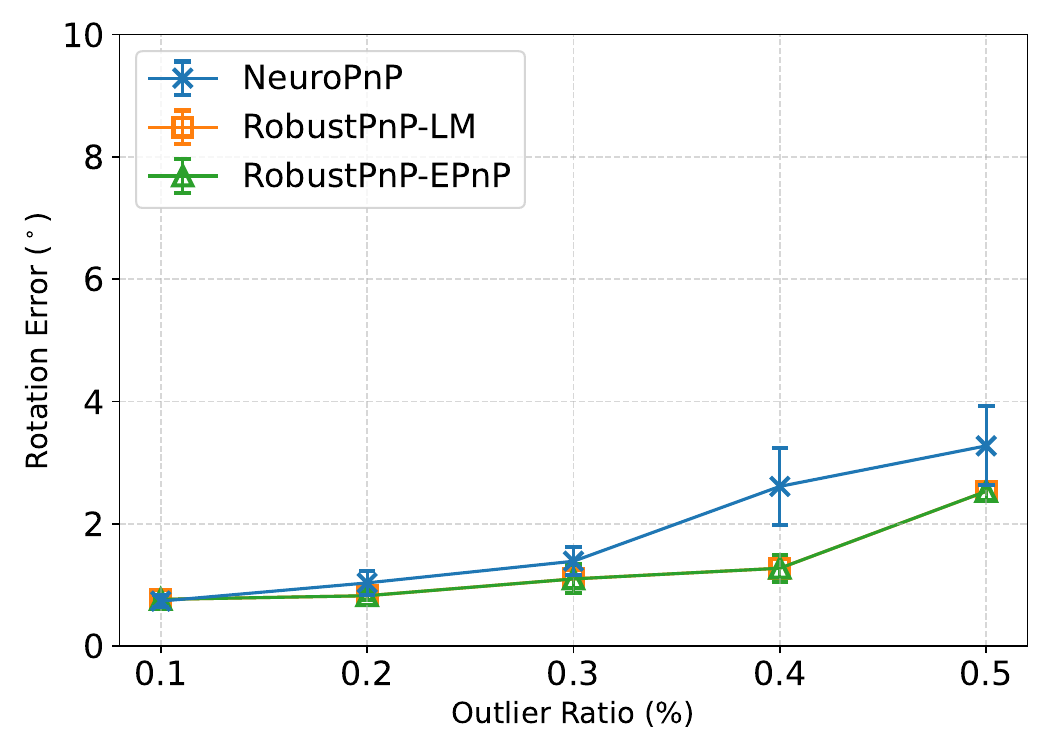}
    \caption{}
    \label{fig:fresh_mean_Rot_err}
\end{subfigure}
\begin{subfigure}[b]{0.32\textwidth}
    \includegraphics[width=0.99\columnwidth]{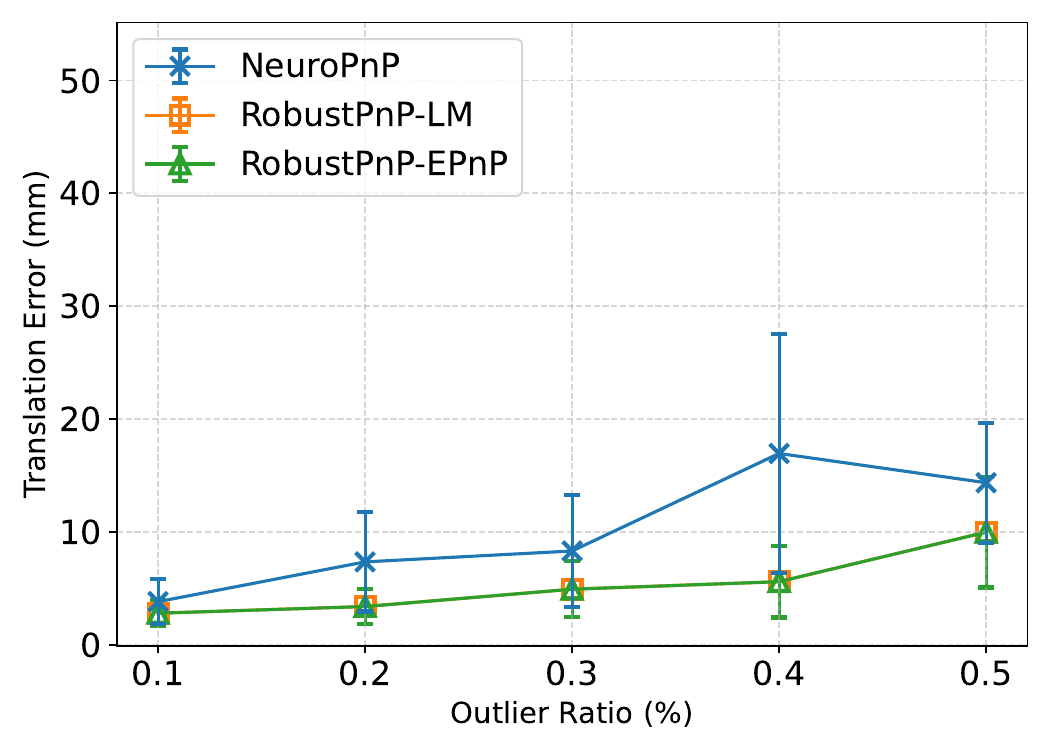}
    \caption{}
    \label{fig:fresh_mean_Trans_err}
\end{subfigure}
\caption{Performance comparison between \neuropnp~and two robust PnP baselines across different outlier ratios on FRESH. AUC values closer to 1 indicate superior inlier detection; lower rotation and translation errors indicate more accurate pose estimation.}
\label{fig:fresh_verify_neuropnp}
\end{figure*}

\paragraph{Validation data}

We used the data generated in Sec.~\ref{sec:datasets}. Since a large number of runs needed to be executed per method, we subsampled 20 PnP instances per object in E-POSE and FRESH. We simulated realistic PnP problems from the ground truth 2D-3D correspondence set in each PnP instance, by perturbing the 2D points with zero-mean Gaussian noise with standard deviation $\sigma_{\text{in}}$, followed by outlier perturbation on $\eta \%$ of points, where $\eta \in \{10,20,30,40,50\}$. Outliers were generated by adding a random displacement of magnitude $(m\cos\phi, m\sin\phi)$, where $\phi \in [0,2\pi)$, $m \in [3,10]$ and $\sigma_{\text{in}} = 1$ for E-POSE, and $m \in [10,50]$ and $\sigma_{\text{in}} = 3$ for FRESH.

\paragraph{Metrics} Inlier/outlier identification accuracy was evaluated using Area Under the Curve (AUC)~\cite{yang2024robust}, there the more accurate the inliers and outliers were recognised, the the higher the AUC value. Accuracy of the final pose was measured via rotation and translation errors~\cite{hodavn2016evaluation}. Results were averaged over 10 runs with std deviations reported.

Note that the runtime comparisons were omitted in this experiment since CPU simulation of NeuroPnP does not reflect the algorithm performance on the intended hardware.

\paragraph{Results}
Figs.~\ref{fig:epose_verify_neuropnp} and~\ref{fig:fresh_verify_neuropnp} plot the AUC, rotation error and translation error across E-POSE and FRESH. All methods degraded as the outlier percentage grew. E-POSE was visibly harder than FRESH, especially in rotation estimation, possibly due to the lower data quality (lower resolution, higher noise). The overall trends of NeuroPnP were similar to the other methods, indicating that it is a competitive robust PnP algorithm. While rotation error of NeuroPnP appeared to be higher in E-POSE, the figures were subject to a high variance for the quality reasons above.

\subsection{Power draw and runtime on Loihi 2}\label{sec:loihi_exp}

\paragraph{Implementation and synthetic data}
We implemented NeuroPnP on Intel Loihi 2. Note that our access to Loihi 2 was conducted remotely via Intel's Neuromorphic Research Community (INRC) program.

While our algorithm scales linearly~\eqref{eq:scaling} and 8192 logical neurons were available, the 8 bit precision in \emph{Synapse} (Sec.~\ref{sec:loihi2intro}) prevented the encoding of realistic data values and a variety of data. To circumvent the constraints, we produced synthetic PnP instances with $N = 15$ integer correspondences. The ground truth pose was ensured to be close to integral as well. Twenty such instances were generated with the number of outliers being 3 to 6, and $\epsilon_{in} = 1$.

\begin{figure*}[t]\centering
\begin{subfigure}[b]{0.32\textwidth}
    \includegraphics[width=0.99\columnwidth]{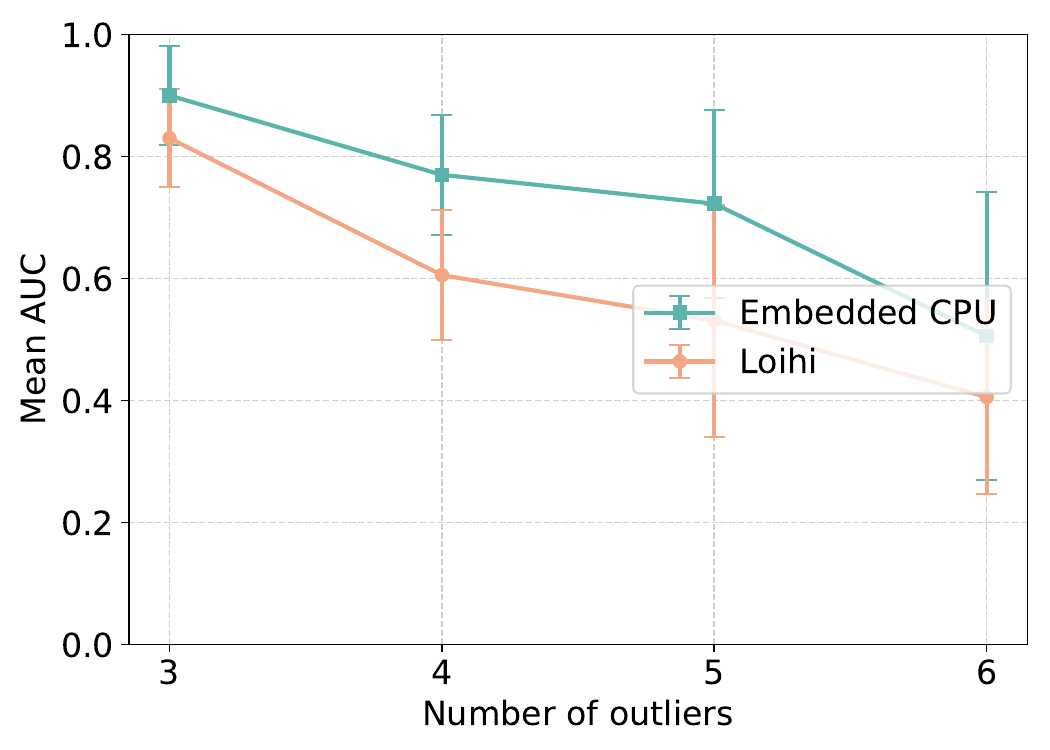}

    \caption{}
    \label{fig:solution_quality_N15}
\end{subfigure}
\begin{subfigure}[b]{0.32\textwidth}
    \includegraphics[width=0.99\columnwidth]{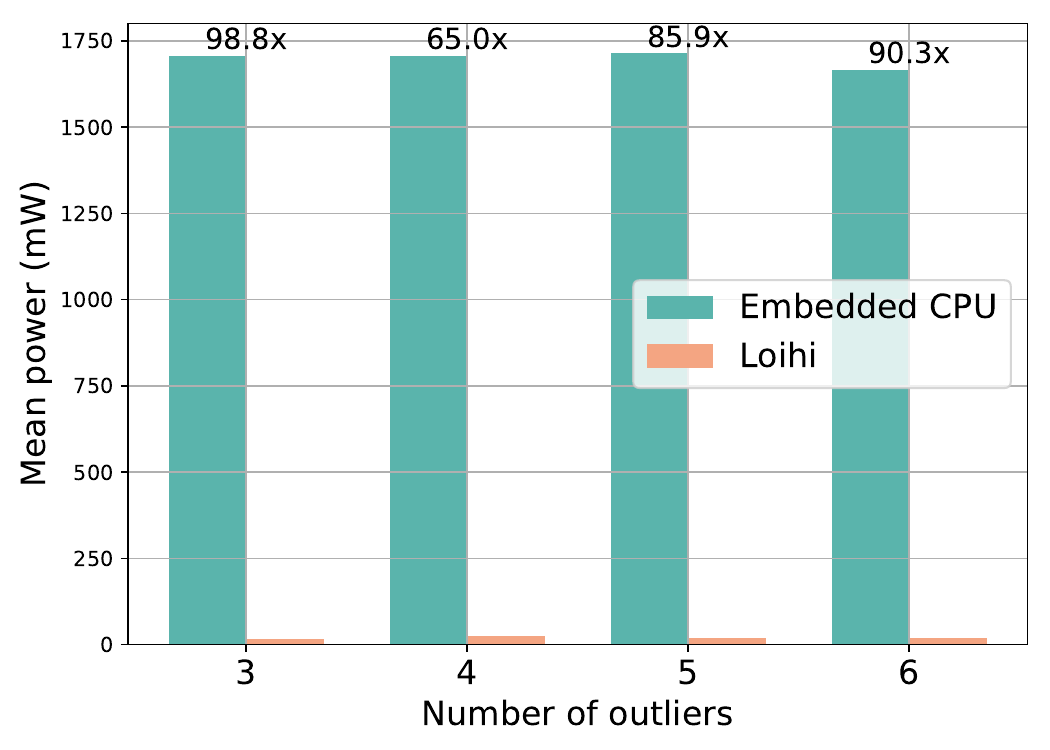}
    \caption{}
    \label{fig:power_consumption_N15}
\end{subfigure}
\begin{subfigure}[b]{0.32\textwidth}
    \includegraphics[width=0.99\columnwidth]{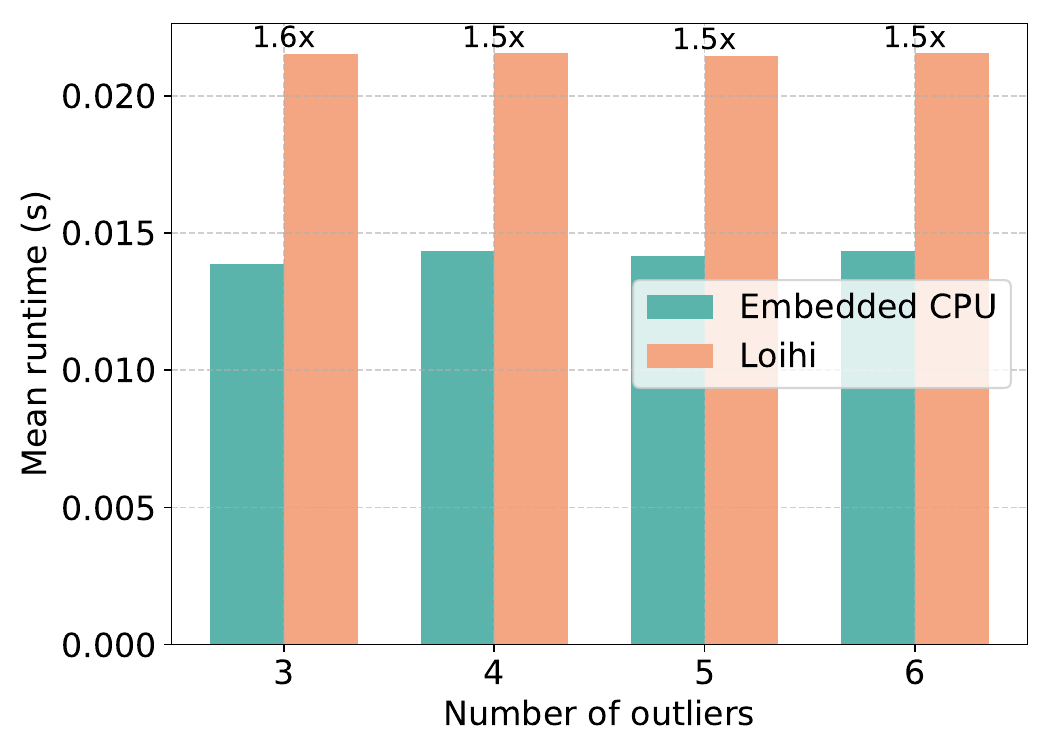}
    \caption{}
    \label{fig:runtime_N15}
\end{subfigure}
\caption{Comparing accuracy, power draw and runtime of \texttt{RobustPnP-EPnP} on an embedded CPU and NeuroPnP on Intel Loihi 2.}
\label{fig:lohiresults}
\end{figure*}

\paragraph{Comparison and metrics}

We compared against \texttt{RobustPnP-EPnP} as executed on a single board computer (SBC) with an Arm Cortex A78AE v8.2 embedded CPU and 16 GB RAM. In addition to the AUC metric used in the previous section, we measured the power draw and runtime of the algorithms, achieved as follows:
\begin{itemize}
    \item For Loihi 2, the onboard Profiler (accessed remotely) was used for both power and runtime recording.
    \item For the embedded CPU system, runtime was obtained with the \textit{time} Python library, while power draw was measured using an external power meter.
\end{itemize}
In both cases, we first recorded the average static power when the system was idle for 5 minutes. Then, we measured the total power $P_{total}$ while executing the workload. Following~\cite{blouw2019benchmarking, mangalore2024neuromorphic, pierro2024solving}, the dynamic power was computed as
\begin{align}
    P_{dynamic} = P_{total} - P_{static}.
\end{align}

\paragraph{Results}

Fig.~\ref{fig:lohiresults} shows the performance values, averaged over 30 trials. The accuracy of NeuroPnP on Loihi 2 was lower than \texttt{RobustPnP-EPnP} on a CPU. However, it should be reminded NeuroPnP was subject to numerical issues due lower precision, usage of fixed-point representation and a reduced instruction set (Sec.~\ref{sec:hardware_details}). In any case, the correctness of NeuroPnP had been verified on CPU.

While the runtime of NeuroPnP on Loihi 2 was slightly higher than \texttt{RobustPnP-EPnP} on a CPU, both sets of runtimes were in the same order of magnitude ($\approx 20$ ms).

Where NeuroPnP clearly shined was power draw; when executed on Loihi 2, NeuroPnP drew only a few miliwatts, which was about 1\% of the power draw of \texttt{RobustPnP-EPnP} on a CPU. Given that
\begin{align}
E_{dynamic} = P_{dynamic} * runtime
\end{align}
and the runtimes of both methods were similar, NeuroPnP was evidently much more energy efficient.

\subsection{NeuroPnP for two-stage neuromorphic OPE}

Here, we demonstrate the applicability of NeuroPnP in a neuromorphic OPE pipeline (Fig.~\ref{fig:wide_figure}). We first trained SPose (Sec.~\ref{sec:landmark_regression}) and HRNet~\cite{sun2019deep} to regress the 2D landmarks for E-POSE and FRESH. Note that HRNet is prevalent in many SOTA OPE pipelines~\cite{wang2023bridging, jin2024advanced}. Further, \snnlandmarkregressor~was trained using the event stacks, while HRNet was trained using the event frame from the chunk.

Both models were trained in PyTorch framework, with \snnlandmarkregressor~implemented using sinabs\footnote{https://github.com/synsense/sinabs}, on an NVIDIA RTX 3090 workstation. To improve generalisation, we applied random rotation and scaling as data augmentations. The HRnet was trained for 10 epochs (batch size 24) with Adam optimizer and a $10^{-3}$ learning rate, which decayed by a factor of 0.1 after the 4th and 8th epochs. \snnlandmarkregressor~was trained for 100 epochs with a batch size of 20 using the same optimizer and learning rate. The learning rate was decayed by a factor of 0.5 at epochs $\{15, 30, 60, 80\}$.

\setlength{\tabcolsep}{3pt}
\begin{table}[t]
\centering
\small 
\begin{tabular}{@{} l l l c c @{}}
\toprule
Dataset & LR & PnP solver & Rot. err. ($\downarrow$)  & Trans. err. ($\downarrow$) \\
\midrule

\multirow{4}{*}{E-POSE} 
    & HRNet & \texttt{RobustPnP} & 9.44 $\pm$ 9.88 & 0.017 $\pm$ 0.108 \\
    & HRNet & \texttt{NeuroPnP}  & 10.19 $\pm$ 11.32 & 0.019 $\pm$ 0.147 \\

    & \snnlandmarkregressor & \texttt{RobustPnP} & 7.96 $\pm$ 9.25 & 0.015 $\pm$ 0.022 \\
    & \snnlandmarkregressor & \texttt{NeuroPnP} & 9.04 $\pm$ 11.13 & 0.017 $\pm$ 0.035 \\

\midrule

\multirow{4}{*}{FRESH} 
    & HRNet & \texttt{RobustPnP} & 1.96 $\pm$ 3.56 & 11.32 $\pm$ 44.55 \\
    & HRNet & \texttt{NeuroPnP}   & 2.08 $\pm$ 6.00 & 12.26 $\pm$ 73.22 \\
    & \snnlandmarkregressor & \texttt{RobustPnP} & 2.14 $\pm$ 5.92 & 14.30 $\pm$ 36.77 \\
    & \snnlandmarkregressor & \texttt{NeuroPnP} & 2.02 $\pm$ 6.52 & 13.31 $\pm$ 29.15 \\

\midrule

\multirow{4}{*}{SPADES} 
    & HRNet & \texttt{RobustPnP} & 1.89 $\pm$ 1.35 & 0.104 $\pm$ 0.115 \\
    & HRNet & \texttt{NeuroPnP} & 2.19 $\pm$ 5.61 & 0.116 $\pm$ 0.146 \\
    & \snnlandmarkregressor & \texttt{RobustPnP} & 0.89 $\pm$ 0.66 & 0.059 $\pm$ 0.067 \\
    & \snnlandmarkregressor & \texttt{NeuroPnP} & 0.93 $\pm$ 2.28 & 0.062 $\pm$ 0.078  \\

\bottomrule
\end{tabular}
\caption{Average rotation error ($^\circ$) and translation error (mm) of different 2D landmark regressors (LR) and PnP solvers.}
\label{tab:ope}
\end{table}

Table~\ref{tab:ope} shows the results of the neuromorphic and classical PnP solvers for the 2D-3D correspondences predicted by \snnlandmarkregressor~and HRNet. The PnP solvers provided similar accuracies, with the major differentiating factor being the landmark regressor used. The results establish the viability of NeuroPnP in a two-stage neuromorphic OPE pipeline.

\section{Conclusions}

We proposed a novel algorithm (NeuroPnP) for robust PnP, proved its viability to be deployed on a real neuromorphic processor, and validated its extremely high energy efficiency when executed on the hardware. When NeuroPnP was employed in a two-stage neuromorphic OPE pipeline, results on real OPE datasets demonstrated the competitiveness of the proposed pipeline against established OPE methods.

\subsection{Weaknesses}

Several weaknesses exist in the current work.

\paragraph{Algebraic formulation}

Since NeuroPnP is based on the algebraic formulation of PnP~\eqref{eq:LS_DLT}, it may require more effort in tuning the inlier threshold $\epsilon_{\text{in}}$. On the other hand, the primary goal or robust PnP is in dichotomising inlier and outlier correspondences, hence, it is arguable that a rough inlier threshold is sufficient for the task.

\paragraph{Hardware limitations}

The evaluation of NeuroPnP (and SPose) on actual neuromorphic hardware is hampered by the serious limitation of current hardware in terms of capacity. Without scaling up the input instances to practical sizes, the level of confidence in NeuroPnP in conducting energy-efficient robust PnP cannot be raised. We are hopeful nonetheless that improved technology is on the horizon~\cite{schuman2022opportunities}.

\addtolength{\textheight}{-3cm}   








\bibliographystyle{IEEEtran}
\bibliography{main}

\begin{IEEEbiography}[{\includegraphics[width=1in,height=1.25in,clip,keepaspectratio]{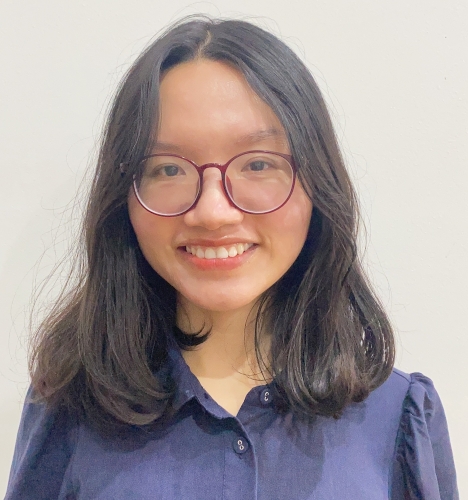}}]{Tam Ngoc-Bang Nguyen} received the B.S. degree in Information Technology from the University of Science, Vietnam National University Ho Chi Minh City, in 2021, and is currently pursuing the Ph.D. degree in Computer Science at Adelaide University. Her research interests include neuromorphic computing and computer vision, with a focus on designing optimization-based vision algorithms on neuromorphic processors. She has published at NeurIPS and an ICCV workshop. \end{IEEEbiography}

\begin{IEEEbiography}[{\includegraphics[width=1in,height=1.25in,clip,keepaspectratio]{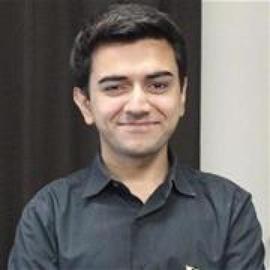}}]{Mohsi Jawaid} received the Bachelor of Computer Science (Advanced) degree from the University of Adelaide, South Australia, in 2018. Following the submission of his Ph.D. thesis at the same institution in 2026, he joined the Australian Institute for Machine Learning (AIML) as a Research Associate. Previously, he worked as a Software Engineer and Data Scientist at Maptek, contributing to machine learning features and specialized 3D spatial visualization tools. He has also served as a Lecturer at the University of Adelaide. His research interests operate at the intersection of artificial intelligence, robotics, and aerospace engineering, with a focus on autonomous systems, multi-robot calibration, vision-based navigation, and domain adaptation for space domain awareness applications. Mr. Jawaid has published in leading robotics and aerospace forums, including IEEE ICRA, IROS, ISPARO, Acta Astronautica, and Aerospace Science and Technology.\end{IEEEbiography}

\begin{IEEEbiography}[{\includegraphics[width=1in,height=1.25in,clip,keepaspectratio]{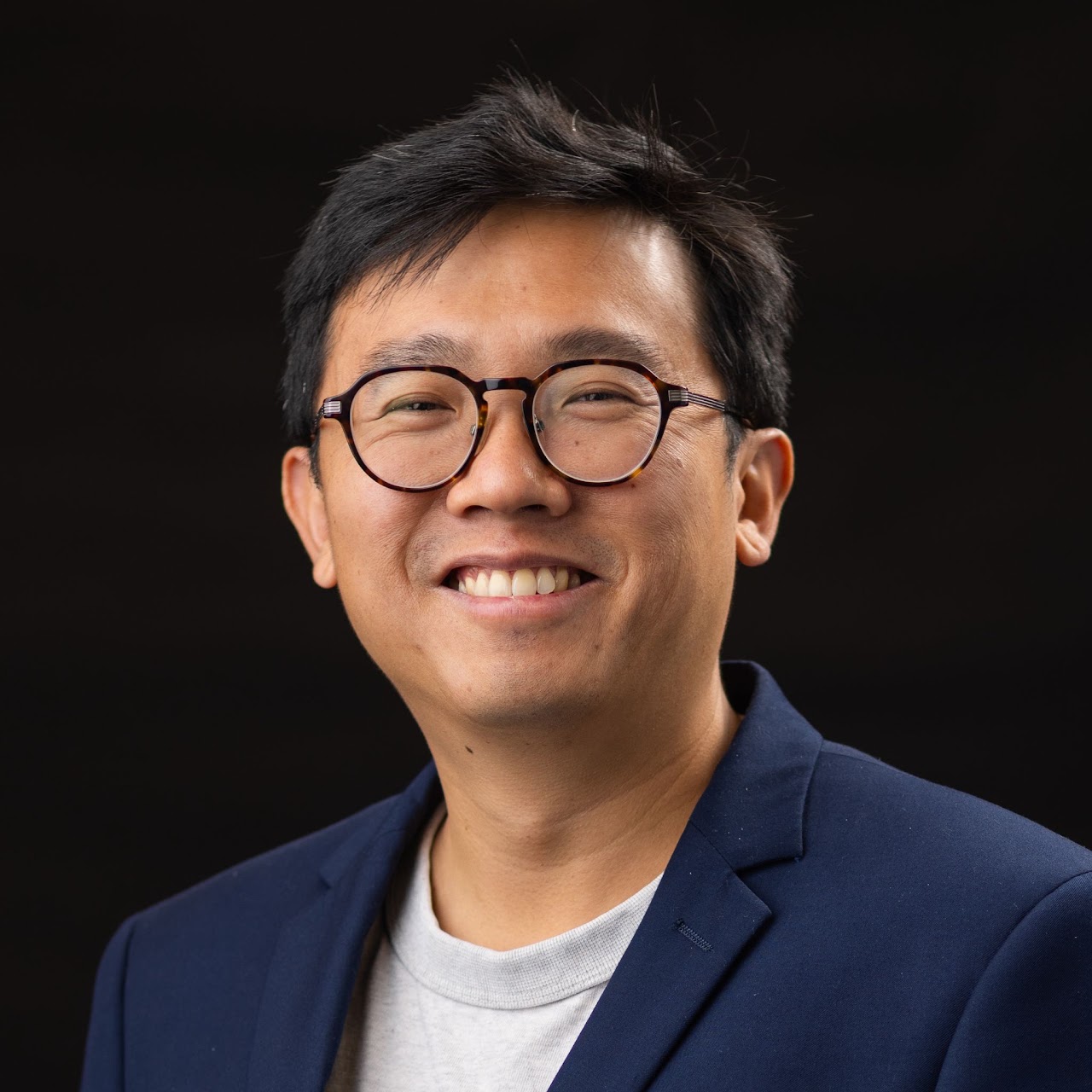}}]{Tat-Jun Chin}
is Professor of Computer Science at Adelaide University, where the leads the AI for Space Group. He received his PhD in Computer Systems Engineering from Monash University in 2007, which was partly supported by the Endeavour Australia-Asia Award, and a Bachelor in Mechatronics Engineering from Universiti Teknologi Malaysia in 2004, where he won the Vice Chancellor’s Award. Tat-Jun’s research interest lies in optimization for computer vision and machine learning, and their application to intelligent satellites and space robotics. He has published more than 150 research articles on the subject, and has won several awards for his research, including a CVPR award (2015), a BMVC award (2018), Best of ECCV (2018), three DST Awards (2015, 2017, 2021), an IAPR Award (2019), an RAL Best Paper Award (2021) and a candidate Best Paper at ECCV (2024). He was a Finalist in the Academic of the Year Category at Australian Space Awards 2021.\end{IEEEbiography}

\end{document}